# RouteRepair: Instance-Level Failure Diagnosis and Targeted Repair in LLM-Based Automated Heuristic Design for Routing Optimization

Binghao Ji[a], Di Huang[a*], Jiahui Fang[a], Zhiyuan Liu[a]

[a] *Jiangsu Key Laboratory of Urban ITS, Jiangsu Province Collaborative Innovation Center of Modern Urban Traffic Technologies, School of Transportation, Southeast University, Nanjing, China, 211189.*

---

**Abstract**

Efficient routing optimization is essential to freight transportation, urban logistics, and shared mobility, where high-quality heuristics are often required under limited computational budgets. Recent large language model (LLM)-based automated heuristic design methods can generate effective routing rules, but aggregate evaluation may mask recurrent failures on particular instance structures. To address this limitation, this study develops RouteRepair, which diagnoses parent-specific weaknesses from instance-level performance and applies targeted modifications to the corresponding heuristic components while protecting behavior that already performs well. Routing evidence, solver behavior, and program context are combined to define bounded repair objectives, and each intervention is validated through matched parent-child evaluation of failure recovery and collateral degradation. Experiments on the traveling salesman problem (TSP) and capacitated vehicle routing problem (CVRP) span constructive search, guided local search, and ant colony optimization. RouteRepair-GLS reduces the mean TSP optimality gap from 1.7476% to 0.7587%, while the constructive CVRP heuristic lowers average route cost by 1.91% relative to the savings heuristic; the generated ACO priors also outperform matched hand-designed priors. These results show that failure-aware, evidence-constrained refinement can improve routing heuristics on difficult instances while preserving performance on cases they already solve well.

---

## 1. Introduction

Routing optimization is central to freight distribution, urban logistics, shared mobility, and emerging autonomous delivery services. The traveling salesman problem (TSP) and the capacitated vehicle routing problem (CVRP) remain two basic models for route sequencing and capacity-constrained distribution. Exact algorithms are indispensable for benchmarking, but their burden grows rapidly with the number of customers and operational constraints. Constructive heuristics, local search, and population-based metaheuristics therefore remain important when good routes must be obtained within limited time (Laporte 2009; Toth and Vigo 2014). Transportation studies have also embedded reinforcement learning and learning-guided decomposition in truck-drone routing, heterogeneous-fleet routing, and large-scale vehicle routing (Bogyrbayeva et al. 2023; Li et al. 2025).

---

* Corresponding author.

*E-mail address:* 1109918035@qq.com(Binghao Ji), dihuang@seu.edu.cn.(Di Huang), fj15252428916@163.com(Jiahui Fang), zhiyuanl@seu.edu.cn(Zhiyuan Liu),

The effectiveness of these methods often depends on compact heuristic components. A constructive procedure must decide which node or customer should be selected next; guided local search must determine which route feature should receive an additional penalty; and population-based methods require rules for variation and selection. Designing these components is difficult because the designer must choose informative state variables, combine competing criteria, and test whether the rule remains useful when customer density, demand dispersion, vehicle capacity, or instance size changes. A rule that works on compact TSP instances may become excessively local on dispersed networks, while a capacity-oriented CVRP rule may consolidate demand too early and create costly detours.

Automated heuristic design (AHD) reframes this task as a search over rules, operators, or executable programs rather than solutions to a single instance. Classical hyper-heuristics and genetic programming established this distinction, although their search spaces generally rely on manually specified primitives (Burke et al. 2013; Dokeroglu et al. 2024). Code-capable large language models (LLMs) provide a broader program prior and support semantically meaningful code variation. FunSearch, EoH, and ReEvo combine LLM generation with execution, evaluation, evolutionary selection, and feedback, turning one-shot generation into iterative program search (Romera-Paredes et al. 2024; Liu et al. 2024; Ye et al. 2024). Most LLM-based AHD methods, however, still rank candidates using a mean objective, average optimality gap, or another scalar fitness. This aggregation hides whether a heuristic is uniformly weak or strong on most instances but repeatedly poor on a smaller, structurally coherent subset.

Recent work addresses heterogeneity through complementary heuristic sets, state-dependent selection, cross-scale training, and robustness-oriented benchmarking (Liu et al. 2025; Yang et al. 2025; Sim et al. 2025). However, these methods do not directly explain how a competitive routing heuristic should be revised when it repeatedly fails on particular spatial, distance, or demand-capacity patterns. RouteRepair addresses this gap by retaining the instance-wise profile of each selected parent and constructing failure, strength, and protection sets. A Failure Diagnosis Expert identifies the likely routing weakness from objective records, instance descriptors, route behavior, and program context, while a separate Targeted Repair Expert modifies only the designated scoring rule or edge-prior function. The repaired heuristic is then evaluated against its parent under matched instances, random seeds, solver settings, and computational budgets. This study makes three contributions:

(i) A new refinement perspective for routing heuristic design. RouteRepair shifts LLM-based automated heuristic design from average-fitness-driven replacement to instance-level repair, enabling recurrent routing failures to be corrected without discarding an otherwise competitive heuristic.

(ii) An evidence-constrained two-expert diagnosis-repair methodology. Parent-specific failure, strength, and protection cases guide a Failure Diagnosis Expert and a separate Targeted Repair Expert, linking routing evidence to bounded code modifications rather than LLM self-assessment.

(iii) A transferable and verifiable repair mechanism. Experiments on five TSP and CVRP adapters show that RouteRepair applies to constructive search, guided local search, and ACO, while improving solution quality and limiting degradation on previously well-solved instances.

## 2. Related Work

### *2.1. Routing Heuristics and Learning-Assisted Routing*

TSP and CVRP have long served as testbeds for constructive heuristics, neighborhood search, and hybrid metaheuristics. The surrounding solver determines the broad search process, but scoring and selection rules often determine how efficiently it uses the available budget. In TSP, these rules include node desirability, edge utility, and neighborhood priorities; in CVRP, they additionally involve demand, residual capacity, route closure, and depot-related costs (Applegate et al. 2006; Laporte 2009; Toth and Vigo 2014; Vidal et al. 2014). Learning-assisted routing generally retains a recognizable optimization structure and learns or adapts one part of the decision process. Transportation applications include joint matching, routing, and pricing for taxi sharing; real-time routing for on-demand buses; reinforcement learning for truck-drone delivery; hierarchical dispatching and routing; and learning-guided decomposition for large vehicle-routing instances (Qiu et al. 2022; Lian et al. 2023; Bogyrbayeva et al. 2023; Si et al. 2024; Kerscher and Minner 2025; Li et al. 2025). These studies differ in problem setting and learning architecture, but they share an important design choice: the learned component is embedded in a clearly specified routing procedure. RouteRepair follows the same principle, while focusing on the diagnosis and revision of executable heuristic rules rather than the training of an end-to-end policy.

### *2.2. LLM-Based Automated Heuristic Design*

Hyper-heuristics search over heuristics or heuristic components rather than directly over solutions (Burke et al. 2013; Qu et al. 2020; Dokeroglu et al. 2024). Evolution through Large Models, Language Model Crossover, and EvoPrompting showed that pretrained code models can act as semantic variation operators over programs (Lehman et al. 2024; Meyerson et al. 2024; Chen et al. 2023). FunSearch established evaluator-guided function search, EoH jointly evolved heuristic ideas and code, and ReEvo introduced comparative reflection as a verbal search signal (Romera-Paredes et al. 2024; Liu et al. 2024; Ye et al. 2024). A unified benchmark further showed that search organization materially affects the quality of LLM-generated heuristics (Zhang et al. 2024).

Subsequent methods have improved search allocation and broadened the design target. MCTS-based AHD and knowledge-guided prompting reuse information from promising candidates (Zheng et al. 2025; Wu et al. 2025). EoH-S searches for complementary heuristic sets (Liu et al. 2025); RedAHD reduces dependence on a manually selected algorithmic framework (Thach et al. 2025); AlphaEvolve evolves larger code regions (Novikov et al. 2025); meta-optimization evolves heuristic optimizers across tasks (Shi et al. 2026); and HeurAgenix combines generation with state-dependent selection (Yang et al. 2025). In transportation routing, Shi and Zhen (2026) use structured prompts, evolutionary search, failure feedback, and repair prompts for vehicle-drone collaborative routing. RouteRepair addresses a complementary problem: diagnosing the instance-specific weakness of a competitive parent and verifying that a localized repair improves targeted cases without eroding established strengths.

### *2.3. Instance Heterogeneity and Verified Repair*

Algorithm-selection and instance-space studies show that performance depends on instance structure (Smith-Miles and Muñoz 2023). Recent benchmarking of LLM-evolved heuristics reaches a similar conclusion: strong results on a design distribution may reflect specialization rather than broad superiority (Sim et al. 2025). Portfolios and dynamic selectors respond by deploying different rules in different regions (Liu et al. 2025; Yang et al. 2025), but they do not explain how the logic of one competitive heuristic should be improved. RouteRepair instead treats diagnosis as an intervention hypothesis. This is related to verbal-reflection agents and automated program repair, where execution feedback is used to revise a previous output (Shinn et al. 2023; Madaan et al. 2023; Le Goues et al. 2012; Monperrus 2018; Olausson et al. 2024). Routing heuristics, however, rarely provide a binary pass-or-fail signal: a program may remain feasible while producing poor routes only on part of the instance set. The repair target must therefore be defined from continuous solution-quality evidence, and collateral effects must be checked where the parent already performs well. Matched parent-child execution, rather than the LLM's own explanation, determines whether the intervention is a genuine repair.

## 3. RouteRepair Methodology

### *3.1. Problem Setting*

Evaluator-guided automated heuristic design represents a heuristic as an executable component embedded in a solver whose feasibility logic and search operations remain externally controlled. This principle underlies

FunSearch, EoH, and ReEvo, although the generated component and feedback mechanism differ across methods (Romera-Paredes et al. 2024; Liu et al. 2024; Ye et al. 2024). It is particularly suitable for routing, where a compact rule can substantially alter the behavior of a mature solver. The editable component may rank the next customer in a constructive procedure, identify an edge to penalize in guided local search (GLS), following Voudouris and Tsang (1999), or provide a static desirability prior to ant colony optimization (ACO).

The methodological challenge arises after a promising parent has been identified. Aggregate fitness separates stronger from weaker candidates but cannot distinguish a uniformly mediocre heuristic from one that is competitive except on a small, structurally coherent subset. RouteRepair retains the outer evolutionary search for broad discovery and adds an inner diagnosis-repair process for strong but locally inconsistent parents.

Let the training set contain N routing instances. Each adapter is associated with a fixed solver skeleton, while the LLM supplies only the designated heuristic component. For the i-th instance, the resulting solution and objective value are

$$y_i(h) = S_k\left(x_i; h\right), \qquad z_i(h) = f_k\left(x_i, y_i(h)\right) \tag{1}$$

All routing objectives considered here are minimized. Let the reference value for the i-th instance be an optimum, a best-known value, or a fixed value computed under the corresponding evaluation protocol. The normalized instance-level loss is

$$r_i(h) = \frac{z_i(h) - z_i^{\mathrm{ref}}}{\max\left(\left|z_i^{\mathrm{ref}}\right|, \varepsilon\right)} \tag{2}$$

In Equations (1) and (2), $x_i$ is routing instance $i$, h is the generated heuristic, and $S_k$ is the fixed solver for adapter $k$. The solver returns the feasible solution $y_i(h)$, while $f_k$ evaluates it to obtain $z_i(h)$. Here, $z_i^{\mathrm{ref}}$ is the reference objective, $\varepsilon>0$ prevents division by zero, and $r_i(h)$ is the normalized excess cost; because the objectives are minimized, a smaller value is better and $r_i(h)=0$ means that the reference value is matched. Rather than discarding these individual outcomes after averaging, RouteRepair retains the complete instance-wise profile.

RouteRepair retains both the instance-wise loss vector and its aggregate mean:

$$\mathbf{r}(h) = \left(r_1(h), r_2(h), \ldots, r_N(h)\right) \tag{3}$$

$$\overline{r}(h) = \frac{1}{N}\sum_{i=1}^{N} r_i(h) \tag{4}$$

In Equations (3) and (4), $\mathbf{r}(h)$ retains the $N$ instance-level losses, whereas $\overline{r}(\mathrm{h})$ is their arithmetic mean. The former is used to locate parent-specific failures, while the latter provides a scalar score for population ranking.

### *3.2. Framework of RouteRepair*

Figure 1 presents RouteRepair as a nested search framework for routing heuristic design. The outer loop generates executable routing heuristics, removes invalid programs, and evaluates the remaining candidates within fixed TSP or CVRP solver backbones. Depending on the adapter, the generated component may rank the next city or customer, select an edge for penalization, or define an edge-desirability prior. Candidate selection uses aggregate route quality, while the complete instance-wise performance profile is retained for subsequent diagnosis. Ordinary variation and optional recombination continue to explore alternative heuristic structures.

The inner loop targets competitive parents that perform well overall but exhibit recurrent weaknesses on particular routing instances, such as elongated customer distributions, sparse node layouts, long-edge patterns, or capacity-stressed demand structures. Failure and strength cases are constructed from the parent's own routing outcomes. The Failure Diagnosis Expert converts route costs, optimality gaps, instance descriptors, solver behavior, and program context into a structured repair hypothesis, while the internal risk review identifies routing behavior that must be preserved. The resulting brief is frozen before the Targeted Repair Expert modifies only the designated scoring rule or edge-prior function. Parent and child are then evaluated under identical routing instances, seeds, solver settings, and computational budgets. A repair is validated only when it improves the diagnosed routing failures without causing excessive degradation on previously well-solved cases. The measured outcome, including unsuccessful repairs, is stored in memory to guide later search.

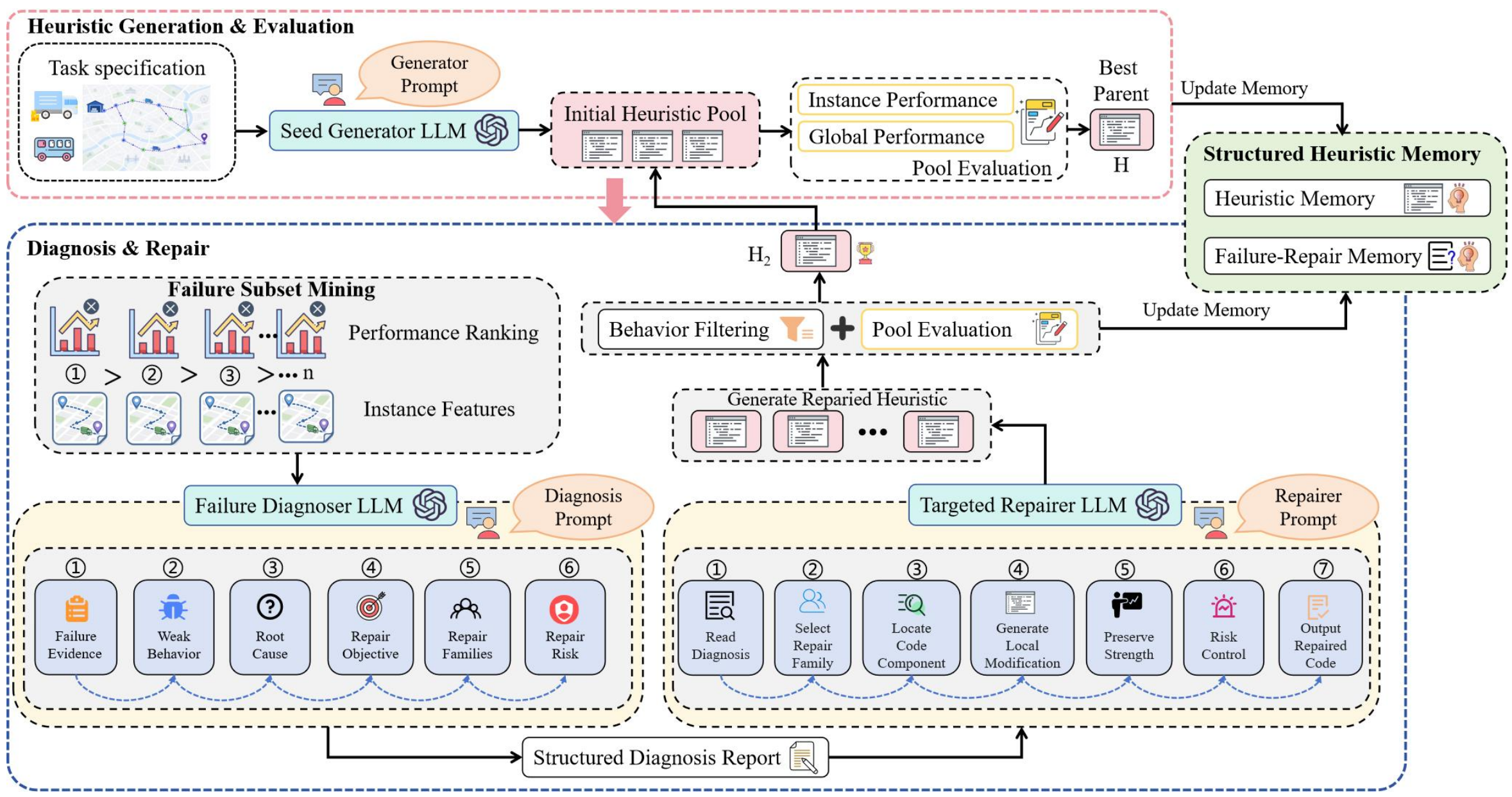

Figure 1: Framework of RouteRepair and its interaction with the outer evolutionary search.

### *3.3. Heuristic Representation and Routing Adapters*

#### *3.3.1. Fixed-Backbone Heuristic Interfaces*

RouteRepair adopts a fixed-backbone representation in which the LLM changes only a designated heuristic interface. The surrounding solver remains responsible for feasibility, state transitions, neighborhood operations, and termination. This separation follows evaluator-guided program search and makes a parent-child comparison meaningful because the solver does not change during repair (Romera-Paredes et al. 2024; Liu et al. 2024). Table 1 lists the five routing adapters and the exact component generated for each backbone.

Table 1: Routing adapters and LLM-generated heuristic components used in RouteRepair.

| Adapter | Fixed backbone | LLM-generated function |
|---|---|---|
| TSP-GLS | Guided local search | score_edge_for_penalty(city_a, city_b, path, distance_matrix, penalties, coords, state) |
| Constructive TSP | Deterministic tour constructor | score_next_node(current_node, candidate_node, unvisited_nodes, distance_matrix, tour_history) |
| Constructive CVRP | Capacity-feasible constructor | score_next_customer(current_node, candidate_customer, unserved_customers, remaining_capacity, distance_matrix, demands, depot) |
| TSP-ACO | Fixed ACO | score_tsp_aco_edges(distance_matrix) |
| CVRP-ACO | Fixed feasible ACO | score_cvrp_aco_edges(distance_matrix, coordinates, demands, vehicle_capacity, depot) |

*3.3.2. Constructive and GLS Adapters*

For the constructive adapters, the generated rule ranks the actions exposed by the fixed solver. At each decision step, the backbone supplies a feasible candidate set together with the state variables permitted by the adapter, and selects the feasible action receiving the highest score. In constructive TSP, the candidate set contains the unvisited cities. In constructive CVRP, capacity-infeasible customers are removed before scoring, so the generated function influences preference but not feasibility.

The TSP-GLS adapter applies the same separation at the improvement stage. The generated function scores an edge already present in the current tour using the edge endpoints, path context, distance matrix, penalties, coordinates, and a restricted solver state. Penalty updates, neighborhood search, and termination remain fixed; only the relative priority assigned to candidate edges is evolved.

*3.3.3. ACO Edge-Prior Adapters*

The ACO adapters change the generated object from a scalar score to a static edge-prior matrix. For a routing instance with an n-by-n distance matrix, the TSP and CVRP adapters return

$$H_h = G_h(D,\gamma) \in \mathbb{R}^{nxn} \tag{5}$$

For TSP, $\gamma$ is empty; for CVRP, it contains coordinates, demands, vehicle capacity, and the depot index. The adapter verifies matrix shape and numerical validity, converts the output to a nonnegative desirability matrix, and sets the diagonal to zero. Following the standard ant colony system construction (Dorigo and Gambardella 1997), the fixed ACO solver combines the prior with pheromone information using the transition weight in Equation (6).

$$w_{ij}^{(t)} = \left[\tau_{ij}^{(t)}\right]^{\alpha} \left[\eta_{h,ij}\right]^{\beta}, \qquad j \in U_t \tag{6}$$

In Equations (5) and (6), $H_h$ is the $n\times n$ static edge-prior matrix generated from the distance matrix $D$ and adapter-specific auxiliary input $\gamma$. The term $\tau_{ij}^{(t)}$ is the pheromone level on edge $(i,j)$, $\eta_{\mathrm{h},ij}$ is the nonnegative desirability derived from $H_{\mathrm{h}}$, $\alpha$ and $\beta$ are fixed exponents, and $U_t$ is the feasible next-node set at construction step $t$.

The generated prior cannot change pheromone evaporation, elite deposits, exponents, ant counts, iteration counts, or random seeds. In CVRP-ACO, visit status and remaining capacity are also hidden from the generated function, so route feasibility remains entirely within the fixed backbone.

## ***3.4. Parent-Specific Case Construction***

RouteRepair does not rely on a fixed list of globally difficult instances, because a case that is difficult for every candidate may reveal little about the selected parent. Instead, the evaluator ranks the parent's own outcomes using an adapter-specific severity measure and normalizes ranks within each problem scale. Within each scale group, rank 1 denotes the worst parent outcome. The normalized within-scale rank is:

$$u_i(h_p) = \frac{m_{g(i)} - rank_{g(i)}\left(i;h_p\right) + 1}{m_{g(i)}} \tag{7}$$

so that the worst instance in each scale receives the largest value. Given prespecified failure and strength fractions, the parent-specific failure set is

$$F(h_p) = Top_{q_f}\{u_i(h_p): x_i \in \mathcal{D}\} \tag{8}$$

whereas representative strength cases are selected from the best-performing part of the same profile,

$$S(h_p) = Bottom_{q_s}\{r_i(h_p): x_i \in \mathcal{D}\} \tag{9}$$

The protection set used for formal collateral-damage evaluation is broader than the small strength subset shown to the LLM:

$$A(h_p) = \mathcal{D} \setminus F(h_p) \tag{10}$$

In Equations (7)-(10), $h_p$ denotes the selected parent heuristic, $g(i)$ identifies the scale group of instance $i$, $m_{g(i)}$ is the number of instances in that group, and $rank_{g(i)}(i;h_p)$ is the parent-specific severity rank. The fractions $q_f$ and $q_s$ determine the sizes of the failure set $F(h_p)$ and strength set $S(h_p)$, while $D$ is the full evaluation set. The set-difference notation $D \setminus F(h_p)$ in Equation (10) means that $A(h_p)$ contains all non-failure instances.

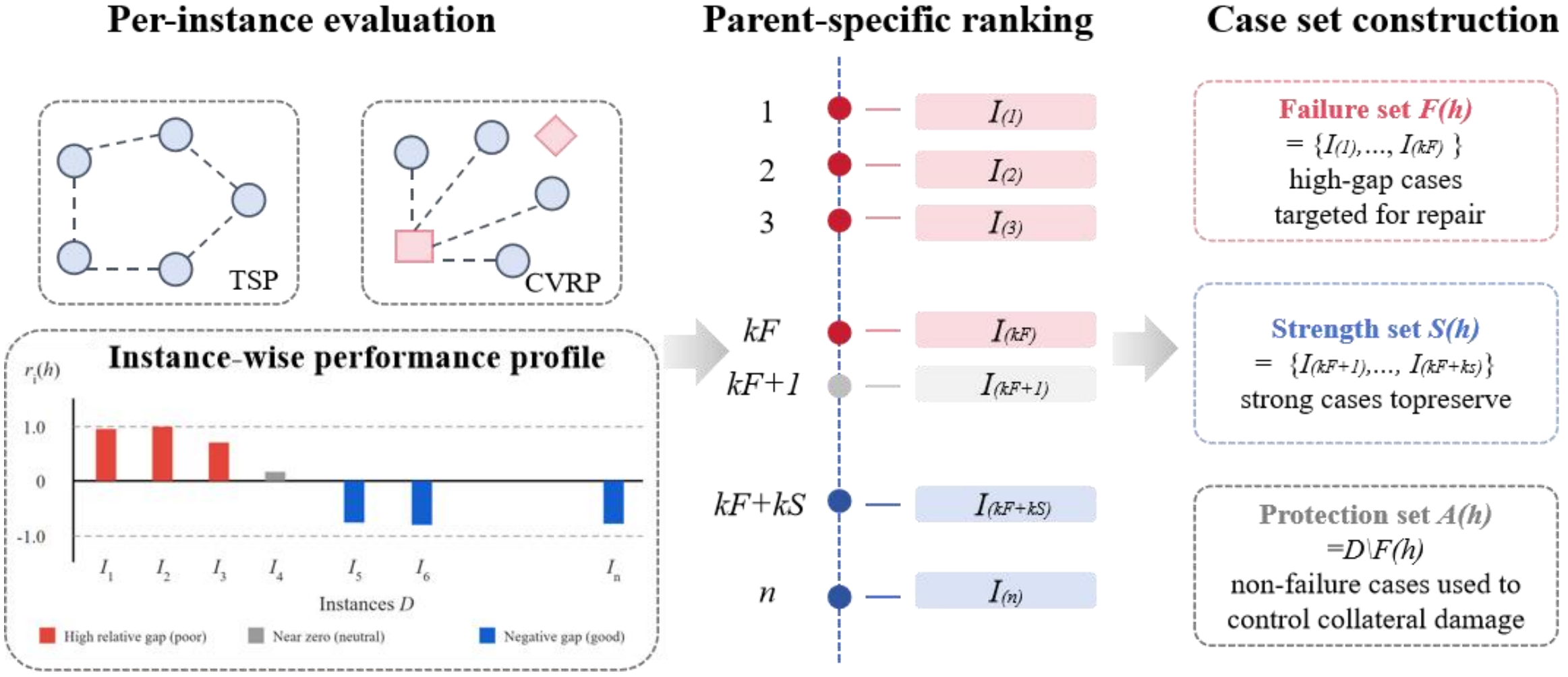

Figure 2: Parent-specific failure, strength, and protection sets.

Figure 2 illustrates the resulting evidence structure. Failure cases define the intervention target, strength cases provide a compact description of behavior worth preserving, and the complete protection set is reserved for the external acceptance test. This parent-specific construction is consistent with the broader observation that algorithm performance can vary substantially across an instance space (Smith-Miles and Muñoz 2023). When a matched baseline is available, the adapter may add contrastive evidence. For example, constructive TSP first highlights cases on which the parent loses to nearest neighbor and then fills the remaining failure slots with high-gap instances; the baseline contrast enriches the diagnosis without replacing the top-tail selection.

The diagnostic packet combines instance-structure descriptors, solution-behavior descriptors recorded by the fixed solver, and statistics of the generated scoring or edge-prior function. Table 2 lists the evidence used by each adapter; all descriptors are deterministic summaries of data already available to the evaluator.

Table 2: Routing-specific evidence used for failure diagnosis.

| Adapter type | Failure-case ranking signals | Diagnostic evidence |
|---|---|---|
| TSP-based adapters | Normalized solution gap within each scale; baseline loss when | Instance structure: size, distance distribution, nearest-neighbor spacing, local/global spacing ratio, density, |

| (TSP-GLS, Constructive TSP, TSP-ACO) | available; best-found tour quality for search-based adapters. | aspect ratio, and node separation. Solution behavior: tour gap, long-edge ratio, crossing pattern, edge-selection preference, stagnation trend, and adapter-specific score/prior distribution. |
|---|---|---|
| CVRP-based adapters (Constructive CVRP, CVRP-ACO) | Normalized routing cost gap; vehicle number; utilization imbalance; capacity pressure indicators. | Instance structure: customer distribution, depot radial characteristics, distance statistics, density, and separation. Demand structure: demand variation, demand-capacity ratio, spatial-demand correlation. Solution behavior: route cost gap, vehicle utilization, route/load variance, fragmentation, late-customer patterns, and adapter-specific score/prior distribution. |

The descriptors distinguish compact and uniform instances from elongated, clustered, or capacity-stressed cases. The local-to-global spacing ratio compares nearest-neighbor and pairwise distances, while the separation proxy identifies locally dense groups connected by relatively long edges. For CVRP, demand-capacity pressure and depot-radial dispersion indicate whether difficult loading decisions are concentrated or geographically scattered.

### *3.5. Failure Diagnosis and Targeted Repair*

#### *3.5.1. Failure Diagnosis Expert*

The Failure Diagnosis Expert receives the parent function, selected failure and strength cases, the evidence in Table 2, the current ranking, and matched-baseline comparisons when available. It does not generate code; it links repeated routing outcomes to the editable component and states a bounded repair objective.

**Prompt for Failure Diagnosis Expert (TSP-GLS Example)**

**[Shared expert instruction]**
Diagnose why the current parent performs poorly on its selected failure cases while identifying the behavior that should be preserved on its strength cases. Treat the root cause as a testable program-level hypothesis. Do not generate Python code.

**[Problem-specific adapter: TSP-GLS]**
Role: diagnose an edge-scoring function used by a fixed GLS solver.
Objective: minimize GLS-improved tour length and reference gap.
Case features: N, tour length, reference length, gap, pairwise-distance statistics, nearest-neighbor statistics, bounding-box aspect ratio, and Standard GLS comparison.

**[Runtime-populated evidence]**
Parent candidate: [PARENT_NAME]
Parent function: [PARENT_CODE]
Aggregate evaluation summary: [EVALUATION_SUMMARY]
Failure cases: [FAILURE_CASES] Strength cases: [STRENGTH_CASES]
Current population ranking: [ROUND_RANKING]
Baseline comparison: [BASELINE_COMPARISON]

**[Shared diagnosis requirements]**
Return six sections: Failure Evidence, Weak Behavior, Root Cause, Repair Objective, Targeted Repair Directions, and Risk Control. Then append exactly one valid JSON block.

**[Required structured diagnosis report]**

```
{
  "failure_evidence": "...",
  "weak_behavior": "...",
  "root_cause": "...",
  "repair_objective": "...",
  "repair_risk": "...",
  "repair_families": [
    {"name": "...", "category": "...",
     "target_failure_pattern": "...",
     "code_change_focus": "...", "why_distinct": "..."}
  ]
}
```

The report must contain exactly five distinct repair families: targeted patch, hybrid recombination, structural rewrite, counterfactual test, and stagnation-escape mutation.

Figure 3: Prompt template of the Failure Diagnosis Expert (TSP-GLS adapter).

The prompt combines a shared expert instruction with an adapter-specific contract covering the function signature, fixed solver boundary, objective, permitted inputs, and failure vocabulary. Figure 3 shows the TSP-GLS version; the ACO variants retain the same diagnostic structure but use the edge-prior contract in Table 1. The diagnosis is treated as a testable program-level hypothesis, not established causality. It must separate observation from inference and identify validated behavior that a broad rewrite could damage; matched parent-child execution, rather than verbal self-assessment, determines whether the proposed intervention is beneficial (Olausson et al. 2024).

*3.5.2. Structured Diagnosis Report and Risk Review*

As shown in Figure 4, the structured diagnosis report is the contract between diagnosis and repair. It records failure_evidence, weak_behavior, root_cause, repair_objective, repair_risk, and one to three distinct repair_families, each with a target failure pattern and code-change focus. The fixed schema converts instance-level evidence into bounded repair directions rather than general commentary or repeated coefficient tuning.

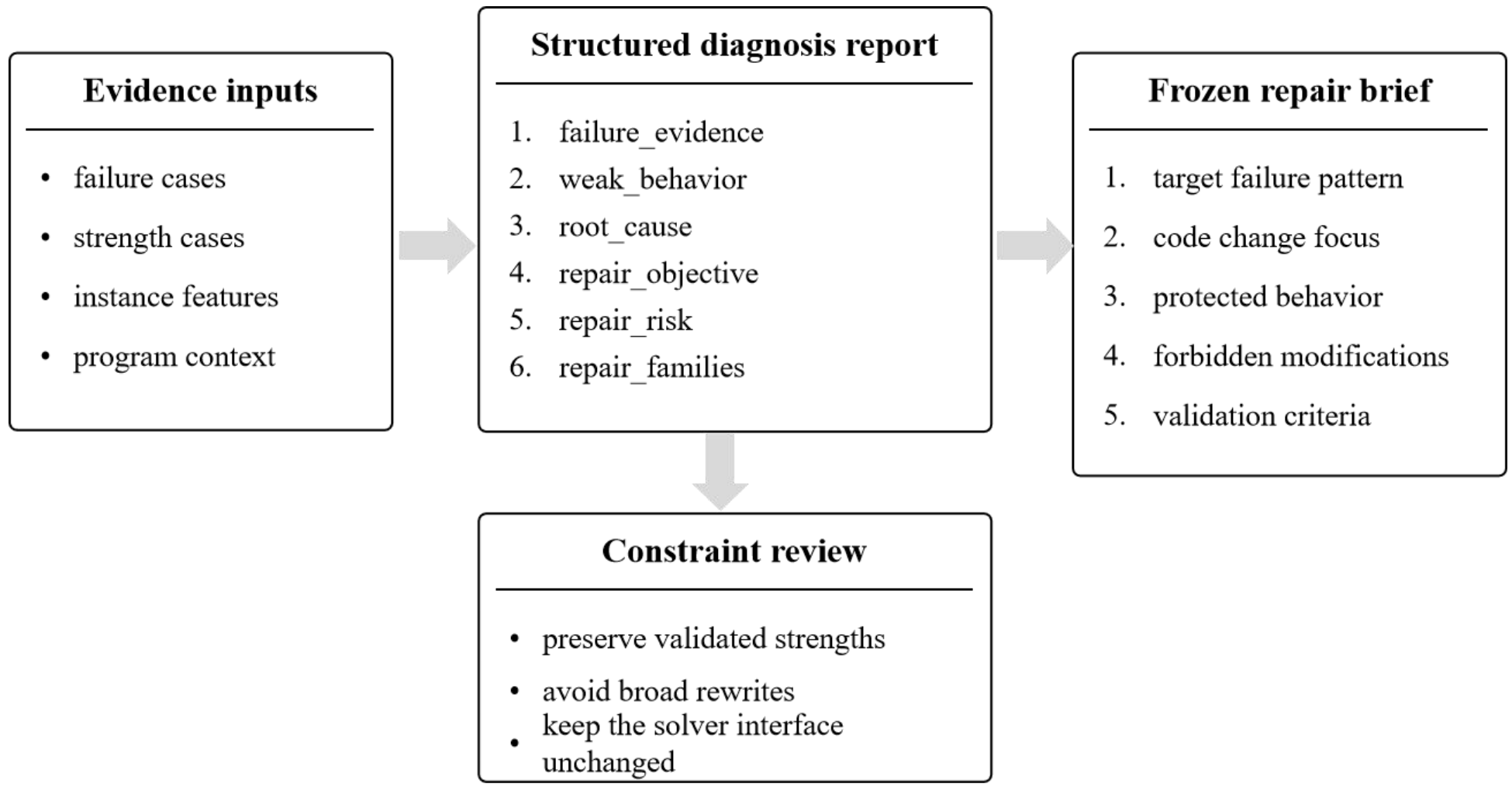

Figure 4: Structured diagnosis report risk review and repair brief to code generation.

Before code generation, an internal review stage examines the diagnosis against the cited evidence, identifies preservation and overfitting risks, removes duplicate repair directions, and produces a concise instruction for the repair stage. This review remains part of the Failure Diagnosis Expert and does not constitute a third code-generating expert. The diagnosis and review are subsequently frozen, preventing the Targeted Repair Expert from redefining the diagnosed weakness after examining a candidate modification.

*3.5.3. Targeted Repair Expert*

The Targeted Repair Expert receives the frozen brief, one repair family, the parent code, compact case summaries, elite context, and a small number of relevant memory records. It must preserve the function signature in Table 1 and make a local or bounded structural change tied to the diagnosed mechanism. GLS and constructive adapters return one finite scalar scoring function, whereas ACO adapters return a deterministic finite edge-prior matrix with the same shape as the distance matrix. Pheromone dynamics, route construction, feasibility checks, and unavailable dynamic state remain outside the editable region. Figure 5 shows the representative repair prompt.

**Prompt for Targeted Repair Expert (TSP-GLS Example)**

**[Shared expert instruction]**
Repair one parent heuristic using the frozen structured diagnosis and reflection/risk review. Preserve useful parent behavior, follow the selected repair family and assigned repair mode, and return one executable function. Repair success is determined only by external evaluation.

**[Problem-specific adapter: TSP-GLS]**
Generated function: score_edge_for_penalty(city_a, city_b, path, distance_matrix, penalties, coords, state).
Return contract: one finite float; a larger value means higher GLS penalty priority.
Allowed signals: distance, penalty, coordinates, local path position, and scalar search state.
Forbidden operations: tour construction, 2-opt, full edge scan, input mutation, randomness, imports, or persistent state.

**[Runtime-populated repair context]**
Primary repair goal: [PRIMARY_GOAL]
Repair parent and selection reason: [REPAIR_PARENT]
Parent code: [PARENT_CODE] Evaluation summary: [EVALUATION_SUMMARY]
Failure cases: [FAILURE_CASES] Strength cases: [STRENGTH_CASES]
Elite context: [ELITE_CONTEXT]
Structured diagnosis: [DIAGNOSIS_JSON]
Reflection/risk review: [REFLECTION_RESULT]
Selected repair family: [SELECTED_REPAIR_FAMILY]
Assigned repair mode: [REPAIR_MODE]
Stagnation summary: [STAGNATION_SUMMARY]
Recent memory: [RECENT_MEMORY]

**[Shared repair requirements]**
The selected repair family is the primary modification direction. Non-patch modes must alter behavior beyond coefficient tuning. Elite mechanisms may inspire the repair but must not be copied verbatim. Use reflection to protect strength cases and memory only as evidence.

**[Required output]**
A brief statement of the intended modification followed by exactly one fenced Python code block containing one repaired function with the original signature and no imports.

Figure 5: Prompt template of the Targeted Repair Expert (TSP-GLS adapter).

### *3.6. Strength-Preserving Matched Validation and Outcome-Aware Memory*

After passing the syntax, interface, and numerical validation gates, each child is compared with its parent under matched conditions. The mean reduction in normalized loss over the parent-specific failure set is:

$$G_F(h_p,h_c)=\frac{1}{|F(h_p)|}\sum_{x_i\in F(h_p)}\left[r_i(h_p)-r_i(h_c)\right] \tag{11}$$

Positive values indicate that the child improves the cases targeted by the diagnosis. To compare interventions across parents with different initial failure severity, RouteRepair uses the recovery rate

$$R_F(h_p,h_c)=\frac{G_F(h_p,h_c)}{\max\left(\frac{1}{|F(h_p)|}\sum_{x_i\in F(h_p)}r_i(h_p),\ \varepsilon\right)} \tag{12}$$

Strength preservation is assessed over the complete non-failure set. Improvements on some protected cases are not allowed to cancel newly created damage on others; collateral degradation is therefore measured by

$$C_A(h_p,h_c)=\frac{1}{|A(h_p)|}\sum_{x_i\in A(h_p)}\max\left(0,r_i(h_c)-r_i(h_p)\right) \tag{13}$$

A child is counted as a successful targeted repair only when

$$Valid(h_c)=1,\qquad R_F(h_p,h_c)\ge\tau_F,\qquad C_A(h_p,h_c)\le\tau_A \tag{14}$$

In Equations (11)-(14), $h_c$ is the repaired child heuristic. $G_F$ is the mean reduction in normalized loss over the failure set, $R_F$ is the corresponding recovery rate relative to the parent's initial failure severity, and $C_A$ is the one-sided average degradation over the protection set; the thresholds $\tau_F$ and $\tau_A$ specify the minimum required recovery and maximum admissible collateral damage, respectively.

Targeted-repair success is deliberately separated from population entry. A child may remain useful to the outer evolutionary search because of a favorable aggregate trade-off, but it is not counted as a verified repair when it fails the recovery or collateral-damage conditions in Equation (14). Each intervention is stored with its diagnosis, repair strategy, parent-child outcomes, recovery, collateral degradation, and verification result. Later prompts retrieve relevant successful and failed cases, allowing memory to reflect measured outcomes rather than unverified advice.

# 4. Experimental Evaluation

The evaluation tests whether instance-level diagnosis and bounded repair improve heuristic components under fixed constructive, local-search, and ACO backbones; whether the mechanism transfers across routing interfaces; and whether gains persist under matched solver settings and explicit LLM budgets. Five TSP and CVRP adapters are examined.

## *4.1. Experimental Settings*

Experiments were implemented in Python 3.10.19 on an HP Victus 16-r0xxx laptop running Windows 11 Home, with an Intel Core i7-13700HX processor, 16 GB RAM, and an NVIDIA GeForce RTX 4060 Laptop GPU. All LLM calls used the DeepSeek API configuration identified as deepseek-v4-flash. The same model served initial generation, diagnosis, internal review, and targeted repair; no task-specific fine-tuning was used. Constructive TSP and CVRP searches used 128 training instances, 25 evolutionary rounds, eight initial candidates, and up to five repair candidates per round. Programs entered evaluation only after syntax, interface, output-shape, and numerical-validity checks. Parent-child comparisons held instances, solver seeds, configurations, and evaluation budgets fixed; adapter-specific ACO settings and held-out sample sizes are reported below.

## *4.2. Traveling Salesperson Problem*

### *4.2.1. TSP-GLS Under a Common Solver Scaffold*

Table 3 isolates the penalty rule by holding the GLS implementation fixed. The comparison includes component-level adaptations of EB-GLS (Shi et al. 2018), GNN-guided local search (Hudson et al. 2022), and Planning of Heuristics (PoH) (Wang et al. 2025). RouteRepair-GLS obtains the lowest mean gap (0.7587%), reducing the gap by 56.6% relative to Standard GLS and by 16.4% relative to the best initial LLM rule. The largest gain occurs on TSP200, where the gap falls from 3.5791% to 1.6897%. The “lite” and “adapted” labels denote component-level transfers into this common scaffold rather than reproductions of the complete source systems.

Table 3: TSP-GLS results under a common solver scaffold (optimality gap, %; lower is better).

| **Method** | **Average** | **TSP20** | **TSP50** | **TSP100** | **TSP200** |
|---|---|---|---|---|---|
| Standard GLS | 1.7476 | 0.0150 | 0.8693 | 2.5271 | 3.5791 |
| Low-penalty GLS | 1.8323 | 0.0690 | 0.9966 | 2.5827 | 3.6807 |
| EBGLS-lite | 1.7578 | 0.0614 | 1.0533 | 2.3650 | 3.5516 |
| GNNGLS-lite | 1.7554 | 0.0407 | 1.0506 | 2.4018 | 3.5285 |
| EoH | 2.2077 | 0.2713 | 1.7217 | 3.0113 | 3.8266 |

| PoH | 1.7066 | 0.0444 | 0.9466 | 2.3557 | 3.4797 |
|---|---|---|---|---|---|
| RouteRepair-GLS | 0.7587 | 0.0358 | 0.3343 | 0.9751 | 1.6897 |

*4.2.2. Budget-Aware Comparison with Complete LLM-AHD Systems*

Table 4 places the same RouteRepair-GLS result in a budget-aware comparison with the complete ReEvo run. RouteRepair attains an average gap of 0.7587% over TSP20-TSP200 using 216 LLM calls, while ReEvo obtains 0.6861% with 302 calls. ReEvo therefore has the lower average gap across TSP20-TSP200, whereas RouteRepair uses 86 fewer LLM calls and records lower gaps on TSP200 and TSP500.

Table 4: Budget-aware comparison with ReEvo runs.

| Method | LLM calls | Average gap (%) | TSP200 gap (%) | TSP500 gap (%) |
|---|---|---|---|---|
| Standard GLS | 0 | 1.7476 | 3.5791 | 4.8924 |
| ReEvo | 302 | 0.6861 | 1.9726 | 2.3439 |
| RouteRepair | 216 | 0.7587 | 1.6897 | 2.2916 |

Compared with the full ReEvo run, RouteRepair requires 86 fewer LLM calls. ReEvo achieves a lower average gap over TSP20-TSP200 (0.6861% versus 0.7587%), while RouteRepair performs better on the larger reported instances: 1.6897% versus 1.9726% on TSP200 and 2.2916% versus 2.3439% on TSP500. These results indicate a quality-cost trade-off rather than uniform dominance, with RouteRepair using fewer LLM calls and remaining competitive as problem size increases. The comparison nevertheless remains specific to the tested implementations, computational settings, and search budgets.

*4.2.3. Constructive TSP and Common 2-Opt Postprocessing*

Table 5 reports constructive gaps before and after a common 2-opt stage. The same 2-opt procedure is applied to every constructor, so the comparison isolates the quality of the evolved starting tour rather than method-specific local search.

Table 5: Constructive TSP results before and after identical 2-opt postprocessing. Values are optimality gaps (%).

| Method | Average | TSP20 | TSP50 | TSP100 | TSP200 |
|---|---|---|---|---|---|
| Nearest Neighbor | 22.5292 | 16.6312 | 23.3894 | 24.0871 | 26.0093 |
| Cheapest Insertion | 16.2027 | 10.5209 | 16.4571 | 18.2242 | 19.6084 |
| Farthest Insertion | 6.0049 | 2.1389 | 5.3997 | 7.2928 | 9.1883 |
| NN + 2-opt | 3.3430 | 1.4812 | 2.8873 | 4.2526 | 4.7507 |
| Cheapest Insertion + 2-opt | 9.0931 | 5.2145 | 8.9455 | 10.6561 | 11.5563 |
| Farthest Insertion + 2-opt | 4.9245 | 1.5299 | 4.1329 | 6.1041 | 7.9312 |
| RouteRepair-Constructive | 16.0760 | 12.2957 | 14.5876 | 17.3112 | 20.1095 |
| RouteRepair-Constructive + 2-opt | 2.8805 | 1.1689 | 2.3272 | 3.3359 | 4.6899 |

Farthest Insertion is the strongest conventional constructor before postprocessing. After identical 2-opt, RouteRepair achieves the lowest mean gap (2.8805%) and the best result at every tested scale; on TSP100, it reduces the gap from 4.2526% for NN + 2-opt to 3.3359%.

Figure 6 reports the scale-wise constructive gaps before and after identical 2-opt postprocessing. The postprocessed results show that the RouteRepair constructor provides stronger starting tours for the common local-improvement stage.

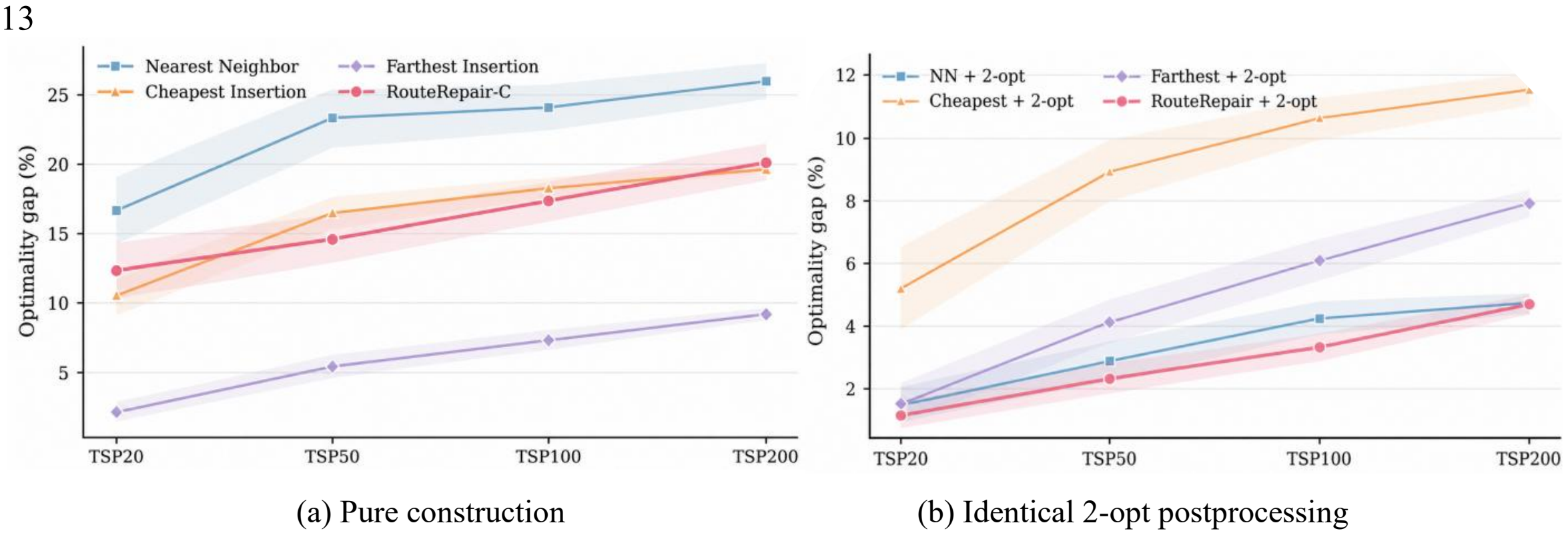

(a) Pure construction (b) Identical 2-opt postprocessing

Figure 6: Constructive TSP performance before and after identical 2-opt postprocessing.

*4.2.4. Transfer to TSPLIB*

Transfer was evaluated without retraining on the complete 30-instance TSPLIB pools. All available instances were retained, including cases on which RouteRepair was not best. Table 6 reports full-pool averages, and Figure 7 gives paired instance-level comparisons. The aggregate comparison confirms that RouteRepair-GLS attains the best full-pool average gap, 1.8742%, improving on Standard GLS (2.0492%) and Low-penalty GLS (2.0505%) while remaining within a comparable runtime range. In the constructive setting, RouteRepair + 2-opt reduces the average gap to 3.9244%, compared with 4.3187% for NN + 2-opt and 7.0208% for Farthest Insertion + 2-opt.

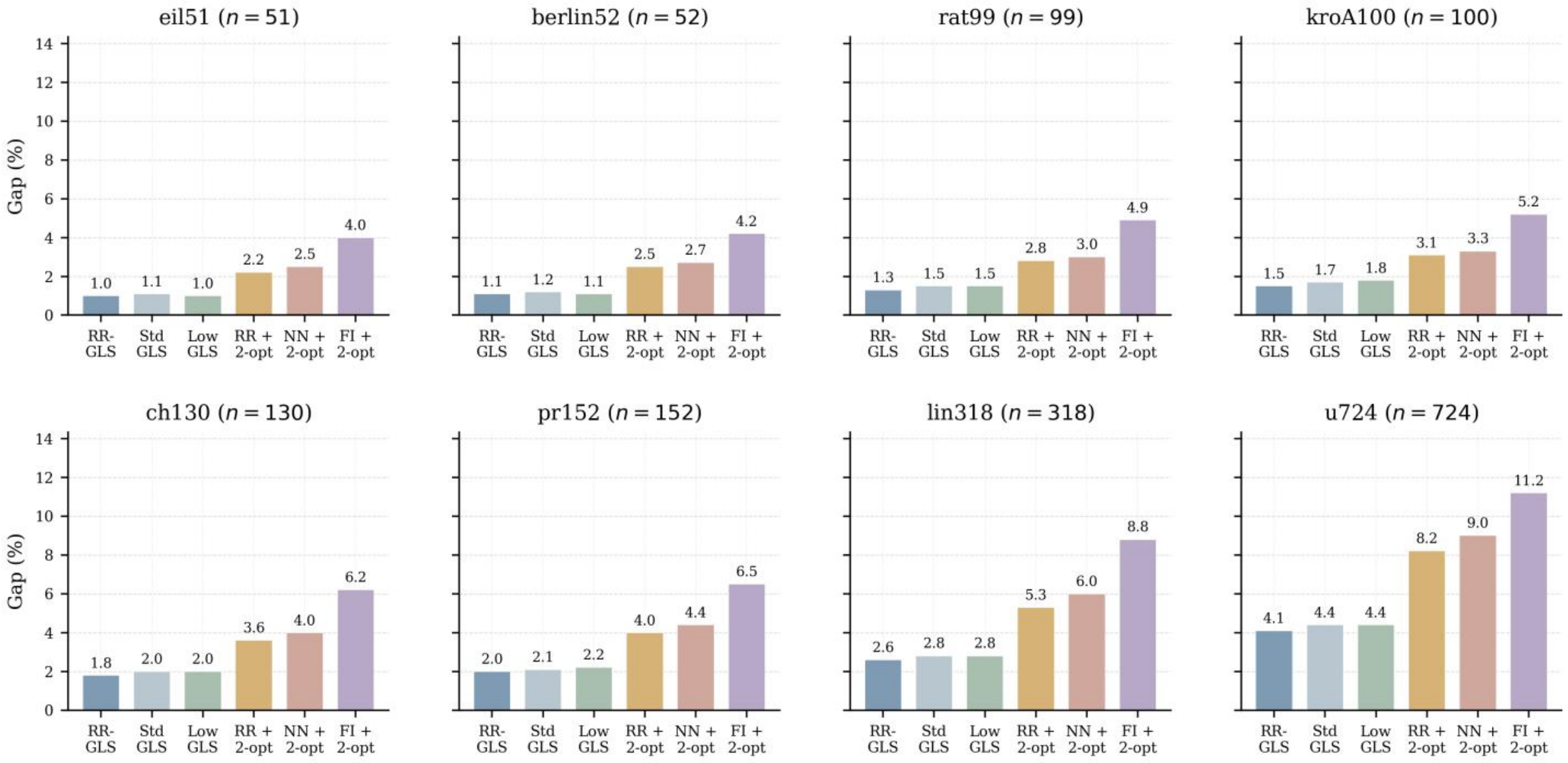

Figure 7: Representative instance-level transfer results on the TSPLIB pool.

Table 6: Aggregate performance on the complete 30-instance TSPLIB pools.

| Task | Method | Instances | Average gap (%) | Runtime (s) |
|---|---|---|---|---|
| GLS | RouteRepair-GLS | 30 | 1.8742 | 18.7226 |
| GLS | Standard GLS | 30 | 2.0492 | 12.0822 |
| GLS | Low-penalty GLS | 30 | 2.0505 | 12.2042 |
| GLS | NN + 2-opt | 30 | 4.5336 | 1.9325 |
| Constructive + 2-opt | RouteRepair + 2-opt | 30 | 3.9244 | 2.6464 |
| Constructive + 2-opt | NN + 2-opt | 30 | 4.3187 | 2.1779 |
| Constructive + 2-opt | Farthest Insertion + 2-opt | 30 | 7.0208 | 1.6881 |

Figure 8 reports paired instance-level transfer results, and Figure 9 shows the best-so-far and population-median training gaps. The paired comparisons indicate that the gains are distributed across multiple instances, while the search trajectories show progressive improvement over successive evolutionary rounds.

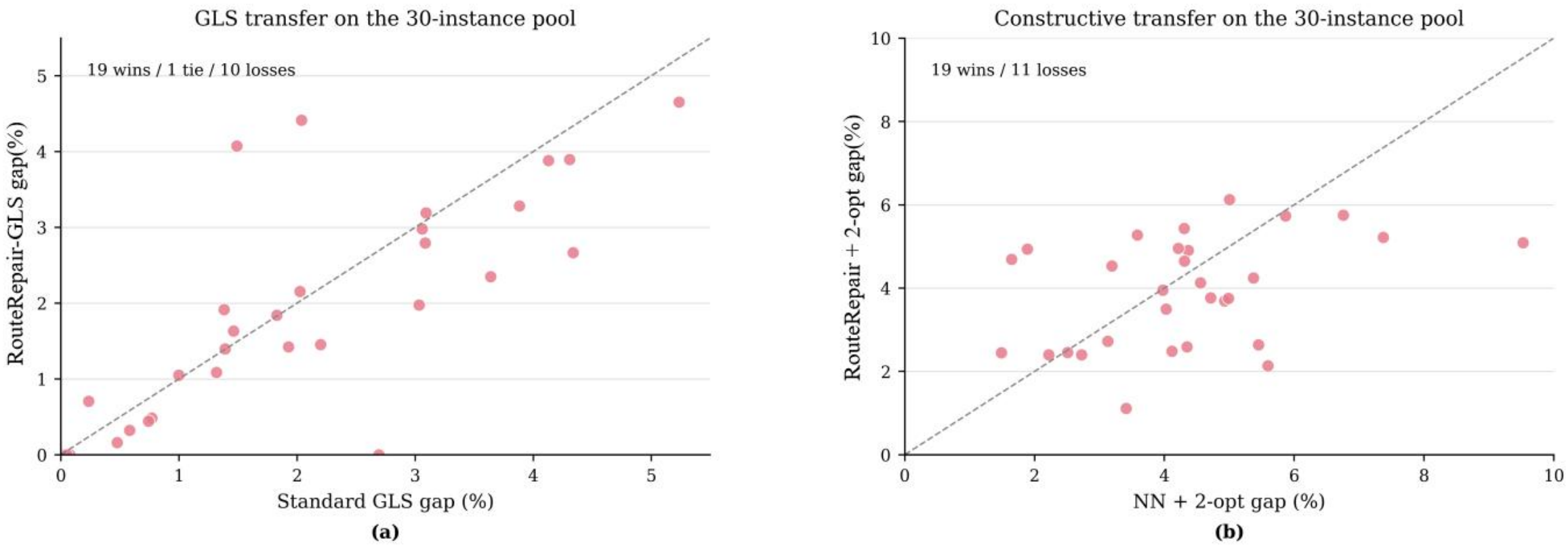

Figure 8: TSPLIB transfer comparison based on paired instance-level performance.

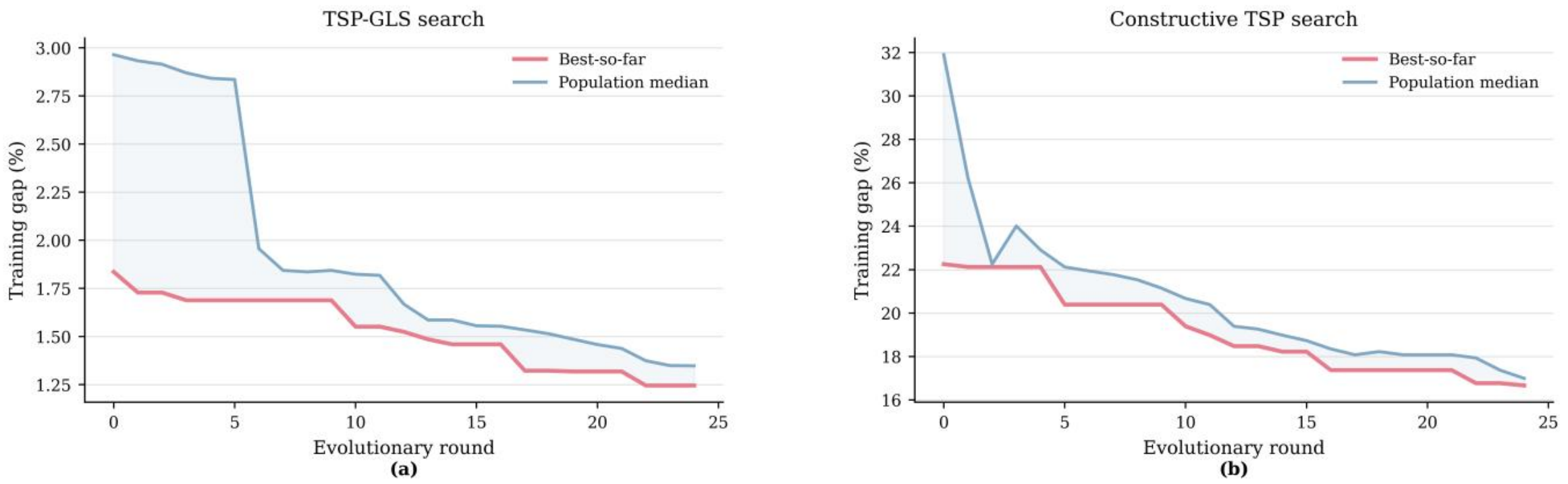

Figure 9: Search dynamics for TSP-GLS and constructive TSP.

#### *4.2.5. TSP-ACO Edge-Prior Generation*

The TSP-ACO adapter generates a static edge-desirability matrix within a fixed ACO implementation. All priors use 24 ants, 40 iterations, and solver seeds 0, 1, and 2. The held-out set contains four instances at each of TSP20, TSP50, TSP100, and TSP200, plus two TSP500 transfer instances. Table 7 reports the matched fixed-prior comparison.

Table 7: TSP-ACO fixed-prior comparison under the common ACO backbone.

| Method | TSP20 (%) | TSP50 | TSP100 | TSP200 | TSP20-200 avg. | TSP500 | 18-case weighted avg. |
|---|---|---|---|---|---|---|---|
| RouteRepair | 0.354 | 1.918 | 7.193 | 11.114 | 5.145 | 19.032 | 6.688 |
| ReEvo | 0.335 | 2.172 | 8.726 | 10.130 | 5.341 | 18.732 | 6.829 |
| Inverse-distance prior | 0.217 | 5.107 | 13.376 | 21.927 | 10.157 | 37.069 | 13.147 |
| Nearest-rank prior | 0.401 | 4.955 | 11.829 | 16.809 | 8.498 | 28.862 | 10.761 |

The scale-wise comparison shows that neither method dominates at every problem size. ReEvo achieves slightly lower gaps on TSP20, TSP200, and TSP500, whereas RouteRepair performs better on TSP50 and TSP100. Nevertheless, RouteRepair achieves the better overall result under the common fixed ACO backbone. Its equal-scale average gap over TSP20-TSP200 is 5.145%, lower than ReEvo's 5.341%, and its weighted average gap across all 18 evaluated instances is also lower, at 6.688% compared with 6.829%. RouteRepair therefore outperforms ReEvo in aggregate while also substantially improving on the inverse-distance and nearest-rank priors.

### *4.3. Capacitated Vehicle Routing Problem*

#### *4.3.1. Constructive CVRP Next-Customer Scoring*

The constructive CVRP experiment replaces only the next-customer scoring rule within a common capacity-feasible constructor; capacity checks, depot returns, state updates, and termination remain fixed. The held-out set contains 64 instances at each of CVRP50, CVRP100, CVRP200, and CVRP500. Table 8 compares RouteRepair with the savings heuristic, adapted EoH and ReEvo guides, and the best initial LLM program, and also reports CVRPLIB transfer.

Table 8: CVRP results under the common constructive next-customer interface.

| Method | Held-out-set average route cost | Gap to OR-Tools Routing GLS (%) | CVRPLIB gap to BKS (%) |
|---|---|---|---|
| Savings heuristic | 27,243.6 | 19.744 | 24.835 |
| EoH-adapted guide | 31,042.2 | 35.531 | 34.434 |
| ReEvo-adapted guide | 27,286.6 | 20.559 | 23.144 |
| Best initial LLM | 28,969.4 | 26.034 | 26.845 |
| RouteRepair | 26,723.6 | 17.523 | 23.001 |

On the held-out test set, RouteRepair attains the lowest average route cost (26,723.6), improving by 1.91% over the savings heuristic, 2.06% over the ReEvo-adapted guide, and 7.75% over the best initial LLM program. Its gap to OR-Tools Routing GLS is also the smallest among the methods restricted to the matched next-customer interface.

On the complete CVRPLIB pool, RouteRepair records a 23.001% gap to the best-known solutions, compared with 23.144% for the ReEvo-adapted guide and 24.835% for savings. The small margin over ReEvo is interpreted as preserved transfer competitiveness rather than a large cross-dataset gain.

Figure 10 includes the four methods for which scale-wise bootstrap results are available. RouteRepair has the lowest mean route cost at CVRP50, CVRP100, CVRP200, and CVRP500. Relative to the ReEvo-adapted guide, the mean reductions are 4.49%, 1.59%, 2.29%, and 1.50%, respectively. Relative to the savings heuristic, the corresponding reductions are 2.36%, 0.69%, 2.52%, and 1.87%. The absolute cost separation is most visible on CVRP200 and CVRP500, while the relative advantage remains positive at every evaluated scale.

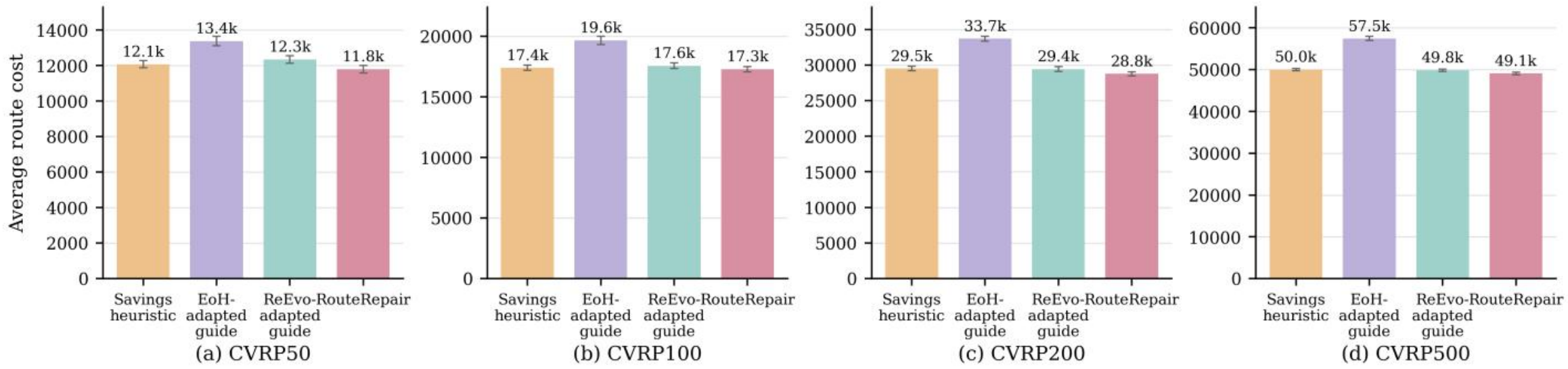

Figure 10: Scale-wise CVRP performance under the common constructive next-customer interface.

The same comparison is summarized in Table 8 at the aggregate level, while OR-Tools Routing GLS is retained only as an external solver reference because it introduces neighborhood search beyond the matched scoring interface.

*4.3.2. CVRP-ACO Capacity-Aware Priors*

The CVRP-ACO adapter modifies only the static edge-prior matrix, while ant construction, pheromone updates, capacity feasibility, route termination, and search parameters remain fixed. RouteRepair is compared with a capacity-aware prior on CVRP50-CVRP500 and with an inverse-distance prior on CVRP1000 under the same ACO backbone. Figure 11 presents the aggregate and scale-wise results, and Table 9 reports the corresponding route costs and reductions.

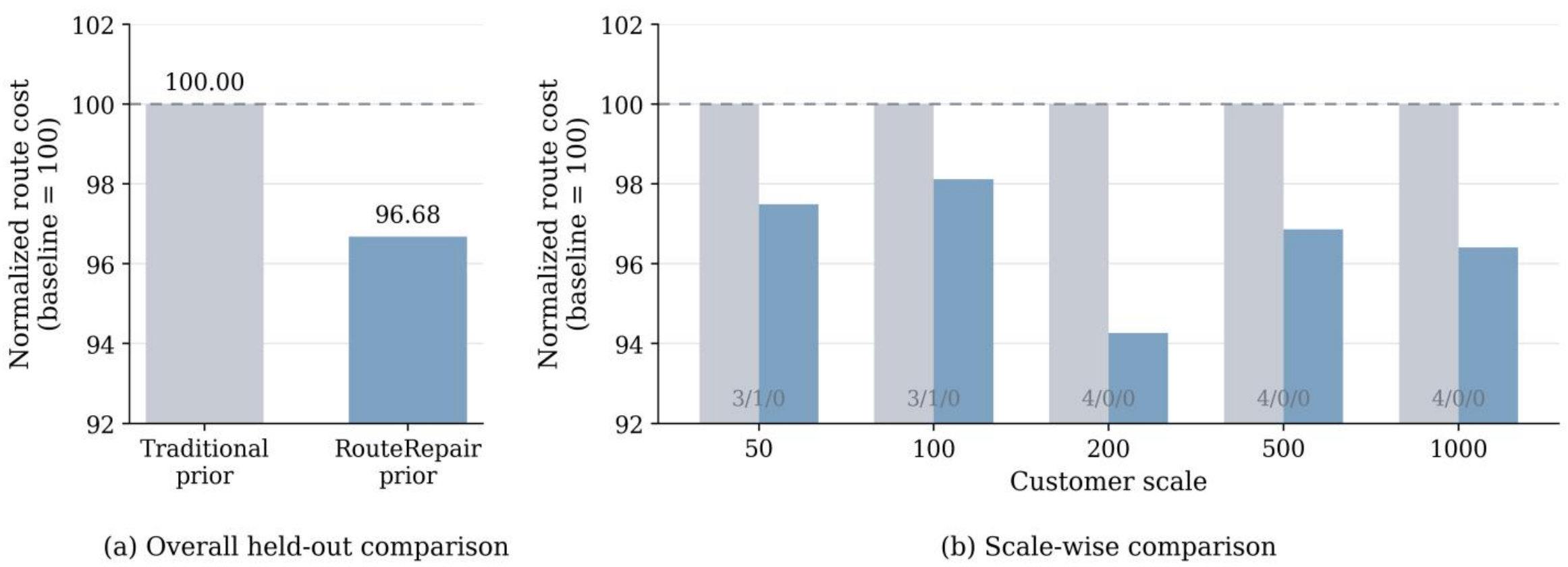

Figure 11: CVRP-ACO comparison with matched hand-designed priors: (a) aggregate held-out result and (b) scale-wise and CVRP1000 generalization results.

RouteRepair lowers the mean route cost at every held-out scale, with reductions ranging from 1.88% on CVRP100 to 5.74% on CVRP200. Across the 16 CVRP50-CVRP500 instances, it improves 14 cases and reduces the overall mean cost from 29,253.0 to 28,184.9, corresponding to a 3.32% reduction. The two losses occur on one CVRP50 instance and one CVRP100 instance, whereas RouteRepair wins all comparisons at CVRP200 and CVRP500.

Table 9: Scale-wise CVRP-ACO comparison with matched hand-designed priors.

| Scale | Instances | Matched prior | RouteRepair cost | Baseline cost | Cost reduction (%) |
|---|---|---|---|---|---|
| 50 | 4 | Capacity-aware | 10,733.5 | 11,018.9 | 2.51 |
| 100 | 4 | Capacity-aware | 16,410.2 | 16,764.1 | 1.88 |
| 200 | 4 | Capacity-aware | 29,970.5 | 31,792.3 | 5.74 |
| 500 | 4 | Capacity-aware | 55,625.6 | 57,436.8 | 3.14 |
| 50-500 | 16 | Capacity-aware | 28,184.9 | 29,253.0 | 3.32 |
| 1000 (gen.) | 4 | Inverse-distance | 100,039.1 | 103,785.8 | 3.59 |

The improvement also transfers to the larger CVRP1000 instances. RouteRepair wins all four comparisons and reduces the mean cost from 103,785.8 to 100,039.1, a reduction of 3.59%. Because vehicle counts and mean capacity utilization remain unchanged, the gains arise from improved edge ordering within the same feasible ACO construction process rather than from additional vehicles or relaxed capacity use. These results show that RouteRepair consistently improves matched hand-designed edge priors under a fixed CVRP-ACO backbone and retains its advantage on larger unseen instances. A direct claim against the official ReEvo CVRP-ACO system is not made because a matched run under the same evaluation protocol was unavailable.

### *4.4. Analysis of Repair Behavior*

#### *4.4.1. Ablation Study*

Table 10 compares full RouteRepair with four ablations: w/o expert-role separation, w/o memory, w/o repair, and w/o failure attribution. All variants use the same test instances, random seed, and evaluation settings; the removed component is the only change in each case.

Table 10: Overall ablation results on TSP-GLS. Lower gaps are better.

| Method | Average gap (%) |
|---|---|
| Full RouteRepair | 0.7587 |
| w/o expert-role separation | 0.9285 |
| w/o memory | 1.7040 |
| w/o repair | 1.7385 |
| w/o failure attribution | 1.7469 |

RouteRepair has the lowest average gap. Merging the expert roles causes a moderate loss, whereas removing memory, targeted repair, or failure attribution raises the gap to 1.70%-1.75%, showing that the main benefit comes from linking parent-specific evidence to bounded repair and reusing measured outcomes.

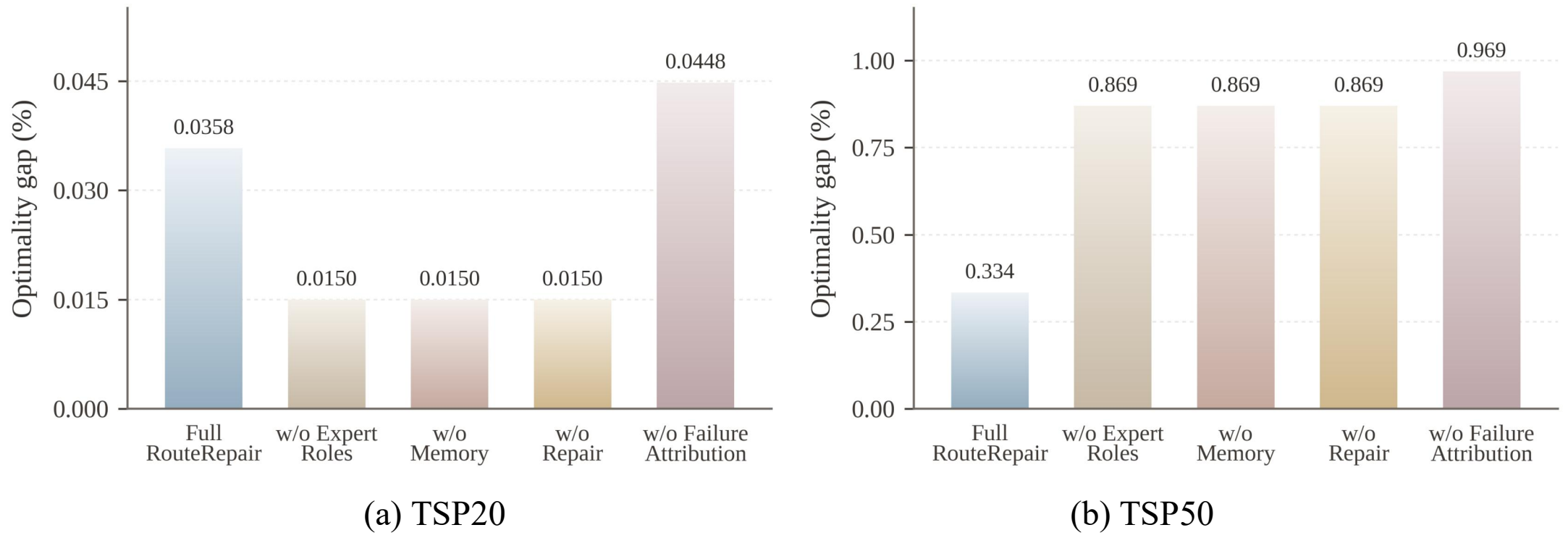

(a) TSP20 (b) TSP50

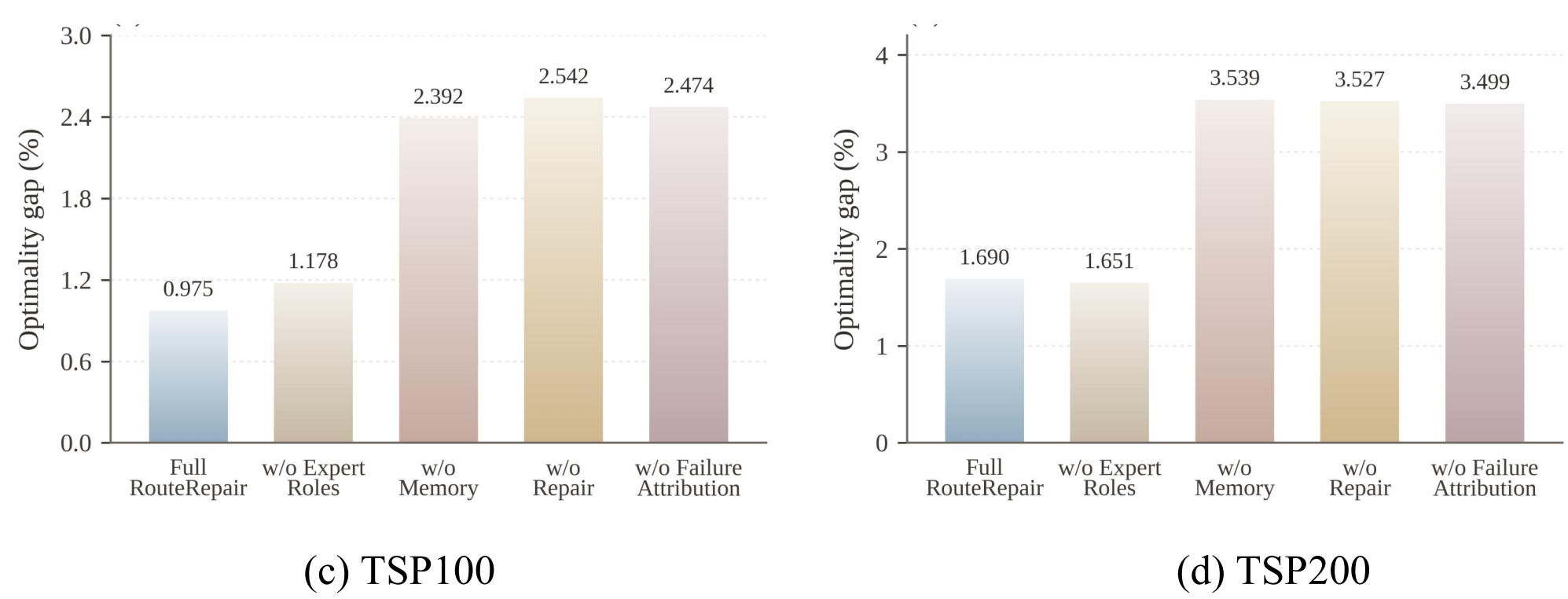

(c) TSP100 (d) TSP200

Figure 12: Ablation results of RouteRepair across different TSP sizes.

Figure 12 reports the results separately for TSP20, TSP50, TSP100, and TSP200. The variants are nearly indistinguishable on TSP20, where all gaps are close to zero. The differences become much clearer on TSP100 and TSP200: removing failure attribution, repair, or memory increases the gap to roughly 2.4%-3.5%. This scale-dependent deterioration suggests that the evidence-to-repair loop becomes more important as problem complexity grows.

### *4.4.2. Instance-Wise Performance Profiles*

Average gaps do not reveal how broadly gains are distributed. Figure 13 therefore normalizes each matched baseline to 100 and orders instances by relative improvement within each scale; values above 100 favor RouteRepair and values below 100 indicate degradation.

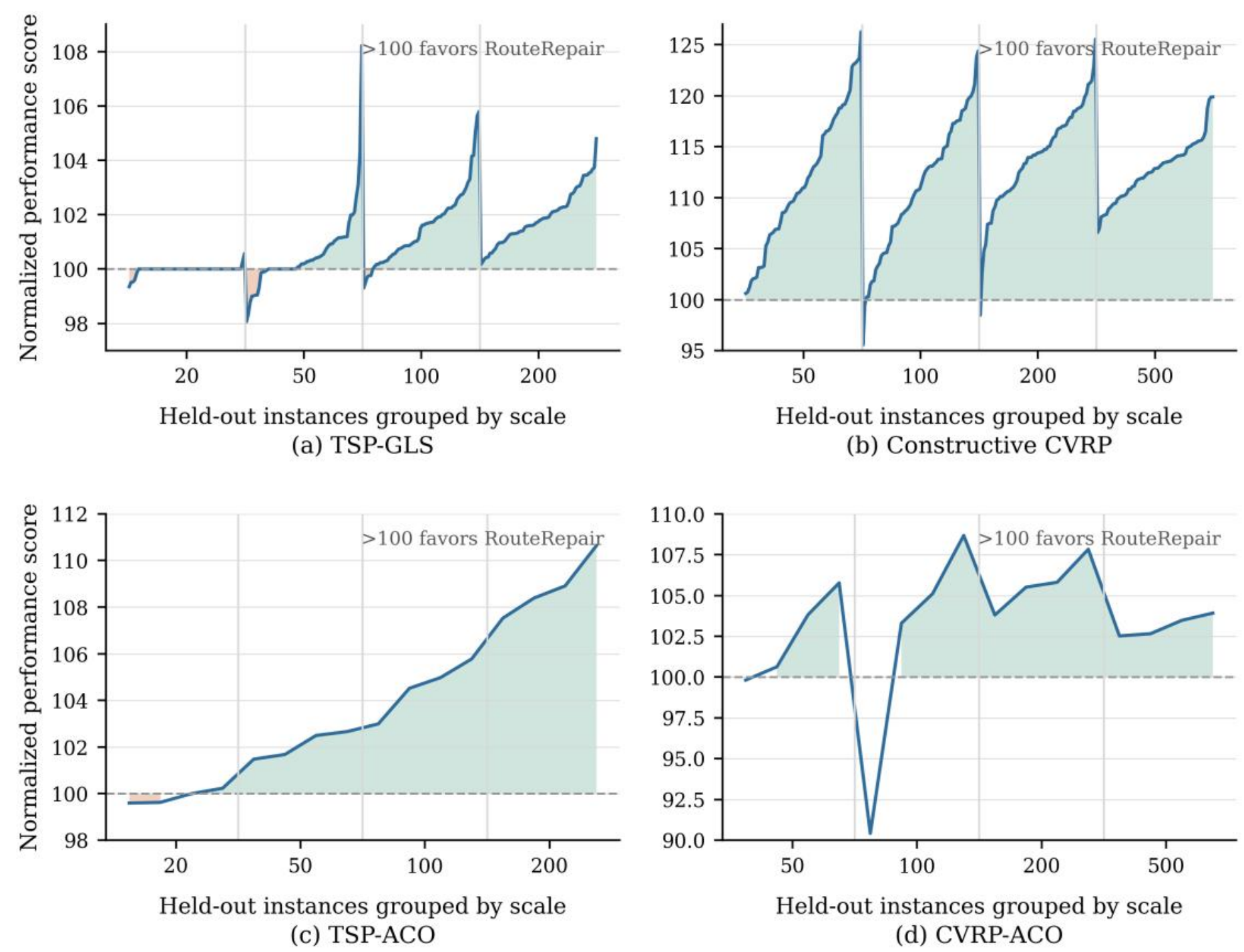

Figure 13: Instance-wise normalized performance profiles across four routing adapters.

The profiles show that TSP-GLS, constructive CVRP, TSP-ACO, and CVRP-ACO place most observations above the matched baseline, although the magnitude of the gain varies across instances and scales. Constructive CVRP exhibits the most consistent positive shift, while both ACO adapters improve most of their completed matched-prior comparisons. These distributions indicate that the repaired scoring and prior functions influence a substantial portion of the test set rather than only a small number of favorable instances.

#### *4.4.3. Failure-Repair Effectiveness*

Table 11 summarizes the instance-level results using a tolerance of 0.1 percentage points around the matched baseline. An instance is classified as improved when its normalized score exceeds 100.1, unchanged when the absolute difference from 100 does not exceed 0.1, and degraded when the score is below 99.9.

Table 11: Instance-level failure-repair effectiveness across routing adapters.

| Adapter | Improved / tested | Improved (%) | Unchanged (%) | Degraded (%) |
|---|---|---|---|---|
| TSP-GLS | 158 / 256 | 61.7 | 30.5 | 7.8 |
| Constructive CVRP | 253 / 256 | 98.8 | 0.0 | 1.2 |
| TSP-ACO | 13 / 16 | 81.2 | 6.2 | 12.5 |
| CVRP-ACO | 14 / 16 | 87.5 | 0.0 | 12.5 |

The results show that RouteRepair improves a majority of the evaluated instances across all four adapters. Constructive CVRP exhibits the most consistent performance, improving 253 of 256 held-out instances, with only three degraded cases. TSP-ACO and CVRP-ACO improve 81.2% and 87.5% of their test instances, respectively, although both retain a small degraded subset. TSP-GLS also produces a clear positive majority, while 30.5% of the instances remain effectively unchanged under the 0.1-point tolerance. Overall, the results indicate that the observed aggregate gains are broadly distributed across the test instances rather than being driven by only a few favorable cases.

## 5. Conclusions

This study addresses a limitation of aggregate-fitness-driven automated heuristic design for routing optimization: a heuristic may achieve strong average performance while repeatedly producing poor routes on a structurally related subset of instances. RouteRepair responds to this problem by preserving the parent's instance-level performance profile, identifying recurrent routing weaknesses, and converting the resulting evidence into a constrained repair objective. The revised rule is then evaluated against its parent under matched instances, solver settings, random seeds, and computational budgets, so that improvement on difficult cases is assessed together with possible degradation elsewhere.

The results show that this instance-level refinement strategy is effective across constructive search, guided local search, and ACO for both TSP and CVRP. RouteRepair-GLS reduces the mean TSP gap from 1.7476% to 0.7587%, while RouteRepair with identical 2-opt postprocessing achieves a 2.8805% constructive TSP gap. For constructive CVRP, the average route cost is reduced by 1.91% relative to the savings heuristic and by 2.06% relative to the ReEvo-adapted guide. The repaired ACO priors also outperform matched hand-designed priors, with CVRP-ACO improving 14 of 16 held-out instances and all four CVRP1000 transfer cases. The ablation results further show that instance-level attribution, targeted repair, and outcome-aware memory become particularly important as routing scale increases.

These findings do not imply that RouteRepair dominates every method or every routing instance. Some differences from ReEvo remain modest or scale-dependent, and the ablation study is based on a single controlled seed. The main contribution is therefore a verifiable refinement methodology that improves competitive routing heuristics without allowing aggregate averages to conceal local failures or collateral losses. Future work should extend the framework to richer vehicle-routing constraints, dynamic and stochastic operations, repeated-seed evaluation, additional LLM families, and adaptive failure-selection criteria.

## References

Applegate, D. L., Bixby, R. E., Chvátal, V. and Cook, W. J. (2006), The Traveling Salesman Problem: A Computational Study. Princeton, NJ: Princeton University Press.

Bogyrbayeva, A., Yoon, T., Ko, H., Lim, S., Yun, H. and Kwon, C. (2023), ‘A deep reinforcement learning approach for solving the traveling salesman problem with drone’. Transportation Research Part C: Emerging Technologies, 148, 103981. https://doi.org/10.1016/j.trc.2022.103981.

Burke, E. K., Gendreau, M., Hyde, M., Kendall, G., Ochoa, G., Özcan, E. and Qu, R. (2013), ‘Hyper-heuristics: A survey of the state of the art’. Journal of the Operational Research Society, 64(12), 1695-1724.

Chen, A., Dohan, D. M. and So, D. R. (2023), ‘EvoPrompting: Language models for code-level neural architecture search’. In Advances in Neural Information Processing Systems, 36.

Dokeroglu, T., Kucukyilmaz, T. and Talbi, E.-G. (2024), ‘Hyper-heuristics: A survey and taxonomy’. Computers & Industrial Engineering, 187, 109815.

Dorigo, M. and Gambardella, L. M. (1997), ‘Ant colony system: A cooperative learning approach to the traveling salesman problem’. IEEE Transactions on Evolutionary Computation, 1(1), 53-66.

Hudson, B., Li, Q., Malencia, M. and Prorok, A. (2022), ‘Graph neural network guided local search for the traveling salesperson problem’. In Proceedings of the International Conference on Learning Representations.

Kerscher, C. and Minner, S. (2025), ‘Decompose-route-improve framework for solving large-scale vehicle routing problems with time windows’. Transportation Research Part E: Logistics and Transportation Review, 204, 104409. https://doi.org/10.1016/j.tre.2025.104409.

Laporte, G. (2009), ‘Fifty years of vehicle routing’. Transportation Science, 43(4), 408-416. https://doi.org/10.1287/trsc.1090.0301.

Le Goues, C., Nguyen, T., Forrest, S. and Weimer, W. (2012), ‘GenProg: A generic method for automatic software repair’. IEEE Transactions on Software Engineering, 38(1), 54-72.

Lehman, J., Gordon, J., Jain, S., Ndousse, K., Yeh, C. and Stanley, K. O. (2024), ‘Evolution through large models’. In Handbook of Evolutionary Machine Learning. Singapore: Springer.

Li, S., Kaisar, E. I., Kwak, D., Hu, B. and Liu, D. (2025), ‘Optimizing heterogeneous capacitated vehicle routing with Linformer and multi-relationship decoding: A deep reinforcement learning approach’. Transportation Research Record: Journal of the Transportation Research Board, 2679(10), 39-60. https://doi.org/10.1177/03611981251338719.

Lian, Y., Lucas, F. and Sörensen, K. (2023), ‘The on-demand bus routing problem with real-time traffic information’. Multimodal Transportation, 2(3), 100093. https://doi.org/10.1016/j.multra.2023.100093.

Liu, F., Tong, X., Yuan, M., Lin, X., Luo, F., Wang, Z., Lu, Z. and Zhang, Q. (2024), ‘Evolution of heuristics: Towards efficient automatic algorithm design using large language model’. In Proceedings of the 41st International Conference on Machine Learning, PMLR 235, 32201–32223.

Liu, F., Liu, Y., Zhang, Q., Tong, X. and Yuan, M. (2025), ‘EoH-S: Evolution of heuristic set using LLMs for automated heuristic design’. arXiv preprint arXiv:2508.03082.

**URL:** *http://arxiv.org/abs/2508.03082*

Madaan, A., Tandon, N., Gupta, P., Hallinan, S., Gao, L., Wiegreffe, S., Alon, U., et al. (2023), ‘Self-Refine: Iterative refinement with self-feedback’. In Advances in Neural Information Processing Systems, 36.

Meyerson, E., Nelson, M. J., Bradley, H., Gaier, A., Moradi, A., Hoover, A. K. and Lehman, J. (2024), ‘Language model crossover: Variation through few-shot prompting’. ACM Transactions on Evolutionary Learning and Optimization, 4(4), Article 27.

Monperrus, M. (2018), ‘Automatic software repair: A bibliography’. ACM Computing Surveys, 51(1), Article 17.

Novikov, A., Vũ, N., Eisenberger, M., Dupont, E., Huang, P.-S., Wagner, A. Z., et al. (2025), ‘AlphaEvolve: A coding agent for scientific and algorithmic discovery’. arXiv preprint arXiv:2506.13131.

**URL:** *http://arxiv.org/abs/2506.13131*

Olausson, T. X., Inala, J. P., Wang, C., Gao, J. and Solar-Lezama, A. (2024), ‘Is self-repair a silver bullet for code generation?’. In Proceedings of the International Conference on Learning Representations.

Qiu, J., Huang, K. and Hawkins, J. (2022), ‘The taxi sharing practices: Matching, routing and pricing methods’. Multimodal Transportation, 1(1), 100003. https://doi.org/10.1016/j.multra.2022.100003.

Qu, R., Kendall, G. and Pillay, N. (2020), ‘The general combinatorial optimization problem: Towards automated algorithm design’. IEEE Computational Intelligence Magazine, 15(2), 14-23.

Romera-Paredes, B., Barekatain, M., Novikov, A., Balog, M., Kumar, M. P., Dupont, E., Ruiz, F. J. R., et al. (2024), ‘Mathematical discoveries from program search with large language models’. Nature, 625, 468-475.

Shi, H. and Zhen, L. (2026), ‘LLM-based automatic heuristic design for vehicle-drone collaborative routing problems’. Transportation Research Part E: Logistics and Transportation Review, 209, 104760. https://doi.org/10.1016/j.tre.2026.104760.

Shi, J., Zhang, Q. and Tsang, E. (2018), ‘EB-GLS: An improved guided local search based on the big valley structure’. Memetic Computing, 10(3), 333-350.

Shi, Y., Zhou, J., Song, W., Bi, J., Wu, Y., Cao, Z. and Zhang, J. (2026), ‘Generalizable heuristic generation through LLMs with meta-optimization’. In Proceedings of the International Conference on Learning Representations.

Shinn, N., Cassano, F., Gopinath, A., Narasimhan, K. R. and Yao, S. (2023), ‘Reflexion: Language agents with verbal reinforcement learning’. In Advances in Neural Information Processing Systems, 36.

Si, J., He, F., Lin, X. and Tang, X. (2024), ‘Vehicle dispatching and routing of on-demand intercity ride-pooling services: A multi-agent hierarchical reinforcement learning approach’. Transportation Research Part E: Logistics and Transportation Review, 186, 103551. https://doi.org/10.1016/j.tre.2024.103551.

Sim, K., Renau, Q. and Hart, E. (2025), ‘Beyond the hype: Benchmarking LLM-evolved heuristics for bin packing’. In Applications of Evolutionary Computation, 386-402. Cham, Switzerland: Springer.

Smith-Miles, K. and Muñoz, M. A. (2023), ‘Instance space analysis for algorithm testing: Methodology and software tools’. ACM Computing Surveys, 55(12), Article 251.

Thach, N., Riahifar, A., Huynh, N. and Chan, H. (2025), 'RedAHD: Reduction-based end-to-end automatic heuristic design with large language models'. arXiv preprint arXiv:2505.20242.

**URL:** *http://arxiv.org/abs/2505.20242*

Toth, P. and Vigo, D. (eds.) (2014), Vehicle Routing: Problems, Methods, and Applications, 2nd ed. Philadelphia, PA: Society for Industrial and Applied Mathematics.

Vidal, T., Crainic, T. G., Gendreau, M. and Prins, C. (2014), 'A unified solution framework for multi-attribute vehicle routing problems'. European Journal of Operational Research, 234(3), 658-673.

Voudouris, C. and Tsang, E. (1999), 'Guided local search and its application to the travelling salesman problem'. European Journal of Operational Research, 113(2), 469-499.

Wang, H., Zhang, X. and Mu, C. (2025), 'Planning of heuristics: Strategic planning on large language models with Monte Carlo tree search for automating heuristic optimization'. arXiv preprint arXiv:2502.11422.

**URL:** *http://arxiv.org/abs/2502.11422*

Wu, X., Wang, D., Wu, C., Wen, L., Miao, C., Xiao, Y. and Zhou, Y. (2025), 'Efficient heuristics generation for solving combinatorial optimization problems using large language models'. In Proceedings of the 31st ACM SIGKDD Conference on Knowledge Discovery and Data Mining.

Yang, X., Zhang, L., Qian, H., Song, L. and Bian, J. (2025), 'HeurAgenix: Leveraging LLMs for solving complex combinatorial optimization challenges'. arXiv preprint arXiv:2506.15196.

**URL:** *http://arxiv.org/abs/2506.15196*

Ye, H., Wang, J., Cao, Z., Berto, F., Hua, C., Kim, H., Park, J. and Song, G. (2024), 'ReEvo: Large language models as hyper-heuristics with reflective evolution'. In Advances in Neural Information Processing Systems, 37.

Zhang, R., Liu, F., Lin, X., Wang, Z., Lu, Z. and Zhang, Q. (2024), 'Understanding the importance of evolutionary search in automated heuristic design with large language models'. In Parallel Problem Solving from Nature - PPSN XVIII.

Zheng, Z., Xie, Z., Wang, Z. and Hooi, B. (2025), 'Monte Carlo tree search for comprehensive exploration in LLM-based automatic heuristic design'. In Proceedings of the International Conference on Machine Learning.